%% file: main.tex
\documentclass[10pt,twocolumn,letterpaper]{article}

\usepackage[pagenumbers]{wacv} 

\input{preamble}

\definecolor{wacvblue}{rgb}{0.21,0.49,0.74}
\usepackage[pagebackref,breaklinks,colorlinks,allcolors=wacvblue]{hyperref}

\def\wacvPaperID{2633} 
\def\confName{WACV}
\def\confYear{2027}

\title{ObjectSplat: Improving Mesh Fidelity and Interactivity for 3D Scenes via Object-Level Mesh Splatting}

\author{
Minhas Kamal, Hiranya Garbha Kumar, Mahedi Kamal, Balakrishnan Prabhakaran\\
State University of New York at Albany, USA\\
{\tt\small \{mxkamal, hgkumar, mkamal3, bprabhakaran\} @albany.edu}
}

\begin{document}
\maketitle

\input{sec/0_01_abstract}
\input{sec/1_00_introduction}

\input{sec/2_00_related_works}

\input{sec/3_00_methodology}

\input{sec/4_00_experiments_results}

\input{sec/5_00_limitations}
\input{sec/8_00_conclusion}


{
    \small
    \bibliographystyle{ieeenat_fullname}
    \bibliography{main}
}

\clearpage

\appendix
\input{sec/9_00_appendix}

\end{document}

%% file: preamble.tex
\usepackage{xcolor} 
\usepackage{makecell} 
\usepackage{multirow} 
\usepackage{soul} 

%% file: sec/0_01_abstract.tex
\begin{abstract}
Splatting-based algorithms reconstruct photorealistic, real-time-renderable, and mesh-exportable 3D scenes from regular images, but they represent a scene as a single monolithic field. Therefore, the reconstruction has no object-level structure, leaving it infeasible for downstream editing or interaction. Moreover, regions that are never directly observed in the input scans are contaminated by the surrounding texture and left uncorrected, capping both mesh fidelity and novel-view synthesis. We propose a decompose-before-reconstruct approach: we segment the instances out of every frame, consider the remaining as background and inpaint it, reconstruct each instance and the background independently with mesh splatting, and compose them into a single scene. Our method significantly improves mesh fidelity (over a 5\% gain in F-score) and novel-view synthesis, while supporting object-wise modifiability and interactivity. The code will be made publicly available.

\end{abstract}

%% file: sec/1_00_introduction.tex
\begin{figure*}[t]
  \centering
  \includegraphics[width=\textwidth]{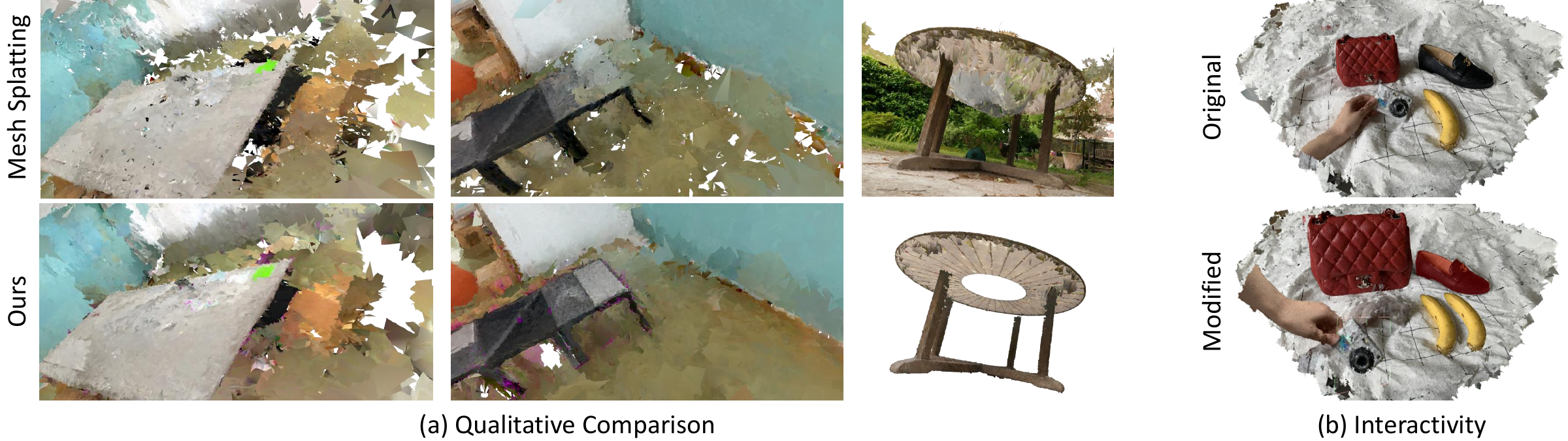}
  \caption{Demonstration of our method's capabilities. \textbf{(a) Qualitative Comparison:} On a novel test view, the baseline holistic reconstruction (top) exhibits significant floating artifacts and texture degradation. In contrast, our object-level approach (bottom) produces a cleaner, more geometrically accurate result. \textbf{(b) Interactivity:} Our object-decomposed representation enables direct manipulation of scene elements, including scaling (purse), appearance editing (shoe), duplication (banana), and rigid transformation (hand). In our pipeline, as backgrounds are inpainted before reconstruction, objects can be moved from their original positions without leaving holes.}
  \label{fig:qualitative_and_interactivity}
\end{figure*}

\section{Introduction}
\label{sec:introduction}

Radiance-field methods have transformed the way photorealistic 3D scenes are reconstructed from 2D scans. Neural Radiance Fields (NeRF) \cite{Mildenhall21NeRF} first demonstrated that a scene could be encoded as a continuous volumetric function and rendered from novel views with excellent realism, but its implicit formulation is slow to train and query. 3D Gaussian Splatting (3DGS) \cite{kerbl233DGS} replaced the neural volume with an explicit set of anisotropic Gaussians and a tile-based rasterizer, achieving comparable quality at real-time framerates. However, 3DGS represents a scene as an unstructured cloud of semi-transparent primitives that float in space and never explicitly model a surface, making its output ill-suited to the polygonal meshes that traditional rendering pipelines and game engines consume. Several subsequent works \cite{guedon2024sugar, yu2024gaussian} try to improve the surface quality of 3DGS. However, 2D Gaussian Splatting (2DGS) \cite{Huang242DGS} takes a direct approach, collapsing primitives onto oriented surface-aligned disks. Following 2DGS, triangle splatting \cite{held25trianglesplatting} and mesh splatting \cite{held25meshsplatting} methods produce explicit meshes that can be imported into 3D engines.

Interactivity - selecting an object and then moving, duplicating, or removing it - requires that a scene be organized into discrete, semantically meaningful objects. This is not merely a structural preference; interaction itself creates new demands. For instance, moving an object reveals its previously unobserved surfaces, making robust novel-view synthesis critical for a seamless user experience. Holistic methods like traditional splatting, however, provide no such object structure. They reconstruct the environment as a single continuous field, a ``soup'' of primitives with no notion of where one object ends and another begins. Consequently, attempting to move an object can cause the holistically-optimized scene to break at the seams, leaving behind corrupted geometry or undefined holes. Although some existing methods attempt to recover object structures after reconstruction, they often produce contact-point artifacts and fail to solve the underlying issue: unobserved areas, now exposed by interaction, are typically filled with noise from the initial joint optimization.

We take a different approach: rather than segmenting a scene after it is reconstructed, we \emph{decompose before we reconstruct}. Our starting observation concerns why holistic reconstruction limits fidelity in the first place. When a scene is fit as a single field, regions that are never directly observed (the underside of a table, the wall behind a cabinet) receive no corrective gradient and are instead filled in by whatever texture surrounds them, leaving stray, mis-colored primitives that degrade the recovered geometry. Reconstructing each object in isolation removes this failure mode: with the background and every other object masked away, the optimizer has nothing to fit but the object itself, and no opportunity to place primitives where the object is not. 


We do a 1:7 train-test split (sparse view) for evaluating our methods ability for novel and unseen view synthesis. Our comparison against vanilla mesh splatting show that per-object decomposition yields significant improvements in mesh fidelity (measured by F-score against ground-truth geometry) together with modest gains in photorealism.
We attribute this to the per-object problem being fundamentally easier: individual objects converge in fewer than half the iterations a full scene requires during training. 
Finally, we study completion as a route to still-higher fidelity, and find that today's obstacle is as much evaluation as method. Our reported metrics exclude every form of completion: diffusion background inpainting and generative amodal completion \cite{wu2025amodal3r} are not view-consistent enough across frames to train on, and even a deterministic flat-surface completion lowers the measured F-score because the ground-truth meshes are themselves incomplete exactly where completion acts, so a recovered surface is scored against a ground truth that lacks it. Assessing completion fairly therefore needs geometry-complete ground truth, such as realistic synthetic scenes, which this setting still lacks. We report this, alongside the consistency limits of automatic multi-view segmentation, to guide future work.

In summary, our contributions are:
\begin{itemize}
  \item A novel method for object-level reconstruction via mesh splatting, built on a decompose-before-reconstruct principle. Our approach incorporates a new optimization strategy that exploits the opacity of primitives to prune extraneous geometry, leading to substantial improvements in mesh fidelity and novel-view synthesis while producing an editable, object-centric scene.
  \item Empirical gains over vanilla mesh splatting in mesh fidelity and novel-view synthesis, with the observation that per-object optimization converges in under half the iterations of a holistic reconstruction.
  \item A systematic analysis of the key bottlenecks in decomposition-based reconstruction. We examine the critical role of cross-view consistency across the segmentation, completion, and inpainting modules, and identify fundamental limitations in current datasets and evaluation metrics that hinder progress.
\end{itemize}

%% file: sec/2_00_related_works.tex
\section{Related Works}
\label{sec:related_works}

\input{tab/related_works_comparison}

\subsection{Monolithic Reconstruction with Splatting}
\label{subsec:rw_meshing}
Neural Radiance Fields \cite{Mildenhall21NeRF} and 3D Gaussian Splatting (3DGS) \cite{kerbl233DGS} established radiance fields as a means of photorealistic, real-time novel-view synthesis, but neither yields the explicit surfaces that downstream graphics pipelines require. Two distinct strategies have since emerged to recover geometry. The first extracts a mesh from a trained 3DGS field as a post-process: SuGaR \cite{guedon2024sugar} regularizes Gaussians toward the surface and meshes them via Poisson reconstruction, Gaussian Opacity Fields \cite{yu2024gaussian} define a level-set opacity field for adaptive surface extraction, and planar- and depth-based variants~\cite{chen2024pgsr, zhang2026rade} further sharpen the recovered geometry. These methods yield accurate geometry, but they optimize shape rather than appearance: the extracted mesh is uncolored, and re-attaching texture from the underlying Gaussians afterward is non-trivial and lossy, which limits the usefulness of the output in rendering and editing pipelines that expect textured assets.

The second strategy makes the surface primitive itself the object of optimization, so that geometry and appearance are learned jointly. 2D Gaussian Splatting (2DGS) \cite{Huang242DGS} collapses each primitive onto an oriented, surface-aligned disk, yielding view-consistent geometry that meshes cleanly, while triangle- and mesh-based splatting \cite{held25trianglesplatting, held25meshsplatting} optimize photorealistic triangle or mesh primitives end-to-end and produce textured, engine-ready meshes directly. Crucially, however, every method in both categories reconstructs the scene as a single, monolithic field with no object-level structure (Table~\ref{tab:comparison}, top block). Our work builds on the second strategy, specifically the mesh-splatting instance.

\subsection{Post-Hoc Segmentation and Shared-Field Editing}
\label{subsec:rw_segmentation}
To make a reconstructed scene editable, a body of work attaches object semantics to the primitives after reconstruction training. One line lifts 2D masks (typically from a promptable segmenter such as SAM \cite{kirillov2023segment}) into the 3D field, either by optimally assigning masks to primitives (FlashSplat \cite{shen2024flashsplat}) or by distilling per-primitive identity and language features during training (Gaussian Grouping \cite{ye2024gaussian}, SAGA \cite{cen2025segment}, Feature-3DGS \cite{zhou2024feature}, LangSplat~\cite{qin2024langsplat}). Because a single primitive straddling an object boundary is shared between neighbours, selecting or editing one object perturbs the other and produces artifacts at contact points; SAGD \cite{hu2026sagd} identifies these boundary primitives explicitly but resolves them only by splitting within the shared field. Building on such segmentations, interactive systems let objects be moved, removed, or simulated (Point'n Move \cite{huang2024point} for click-based manipulation, VR-GS \cite{jiang2024vr} for physics-driven deformation, and text- or physics-based editors~\cite{chen2024gaussianeditor, xie2024physgaussian}), yet each operates on a shared 3DGS field, so moving an object exposes background regions the reconstruction never modelled, which are then filled per edit at inference time rather than reconstructed once (Table~\ref{tab:comparison}, second block). Almost all of this work targets 3DGS, whose primitives lack the mesh accuracy of the 2D-splatting representations above.

Closest to our setting are two concurrent, unpublished preprints. Object-centric 2DGS \cite{rogge2025object} reconstructs a single isolated object on the mesh-accurate 2D-splatting representation, but removes---rather than completes---its background and handles only one object at a time. Split\&Splat \cite{monchieri2026split} instead segments a scene into multiple objects, reconstructs each independently, and merges them, making it the most similar in spirit to our pipeline; however, it operates on 3DGS, leaves the background exposed by removed objects unfilled, and does not target mesh fidelity or the sparse-view regime. Neither, moreover, provides a public implementation, precluding a direct experimental comparison. Our method differs from both by reconstructing every object and a completed background as independent mesh splats, and by doing so under sparse views.

\subsection{Inpainting and Amodal Completion}
\label{subsec:rw_inpainting}
A related line studies how to fill the regions revealed when an object is removed. In the NeRF setting, SPIn-NeRF~\cite{mirzaei2023spin} established the mask-and-inpaint paradigm, supervising the masked region with a 2D inpainter. Gaussian-domain methods extend this to splats: GScream \cite{wang2024learning} regularizes the geometry and features of the filled region for cross-view consistency, InFusion \cite{liu2026infusion} learns depth completion from a diffusion prior to place new primitives at correct depths, and RefFusion \cite{mirzaei2024reffusion} adapts a diffusion inpainter to a reference view and distills it into the 3D field. These methods fill the hole left by a single removal at edit time; they do not produce a persistent, fully completed background that supports arbitrary rearrangement of many objects, which is what our pipeline reconstructs up front.

\begin{figure*}[t]
  \centering
  \includegraphics[width=0.95\textwidth]{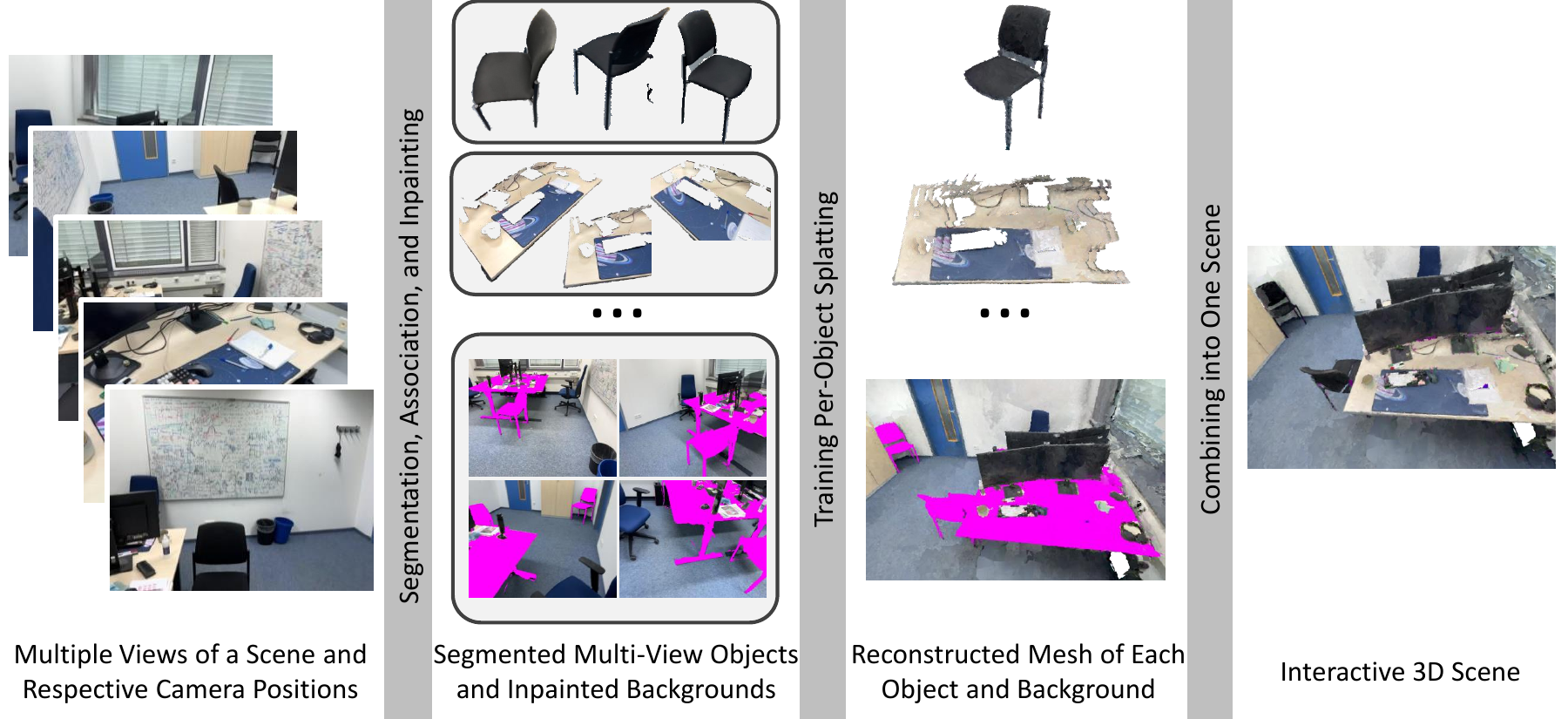}
  \caption{Overview of our pipeline. Every pixel of every posed frame is assigned to exactly one component: an object or the background. Objects and the background, which for our qualitative results is inpainted but is excluded from quantitative evaluation (see Sec.~\ref{subsec:interactivity}), are reconstructed independently; the components are then composed by union into a single, separable scene.}
  \label{fig:pipeline}
\end{figure*}

A complementary question is how to recover the parts of an object that are themselves occluded. Amodal completion methods such as Amodal3R \cite{wu2025amodal3r} hallucinate an object's hidden geometry from its visible extent and could, in principle, further improve the fidelity and completeness of each per-object reconstruction. In practice, however, we find that current amodal methods are not sufficiently consistent across views: injecting their inconsistent completions into the per-object optimization degrades rather than improves it. We therefore treat object completion as a promising but not-yet-reliable direction rather than a component of our pipeline.

\subsection{Sparse-View Radiance-Field Reconstruction}
\label{subsec:rw_sparse}
Because dense captures are impractical in many settings, a growing body of work reconstructs radiance fields from sparse views. Regularization-based methods add geometric priors to few-shot 3DGS (FSGS \cite{zhu2024fsgs} densifies Gaussians guided by proximity and monocular depth, and DNGaussian~\cite{li2024dngaussian} normalizes depth at global and local scales), while pose-free and feed-forward approaches such as InstantSplat~\cite{fan2024instantsplat} and MVSplat~\cite{chen2024mvsplat} initialize geometry directly from stereo or cost-volume priors, and diffusion-prior methods such as ReconFusion~\cite{wu2024reconfusion} synthesize pseudo-observations at unseen viewpoints. Most of these optimize novel-view RGB quality on a monolithic scene; MAtCha \cite{guedon2025matcha} is a notable exception in targeting mesh-quality geometry from few views, but it too reconstructs the scene as a whole. None pairs sparse-view robustness with per-object, mesh-accurate decomposition, the regime in which we report our largest gains (Table~\ref{tab:comparison}, bottom block).

Across these four lines of work, the same gap recurs. Methods that achieve high mesh fidelity reconstruct the scene as a single monolithic field; methods that recover object structure segment an already-reconstructed field, inheriting contact-point artifacts and disocclusion holes that are filled per edit, almost always on 3DGS rather than a mesh-accurate representation, and, because the underlying reconstruction is already fixed, unable to improve its mesh fidelity; inpainting and amodal completion address those holes only one edit at a time, or not yet reliably; and sparse-view methods improve few-shot reconstruction but leave the scene holistic and appearance-oriented. Consequently, no prior method in Table~\ref{tab:comparison} satisfies all of these axes at once. We close this gap by inverting the usual order: instead of reconstructing a scene and then decomposing it, we decompose the multi-view observations first and reconstruct each object, alongside a completed background, as independent mesh splats that merge into a single interactive scene. We detail this method in Section~\ref{sec:methodology}.

%% file: tab/related_works_comparison.tex
\providecommand{\cmark}{\checkmark}
\providecommand{\pmark}{\ensuremath{\sim}}
\providecommand{\xmark}{\ensuremath{\times}}

\newcommand{\compRotHead}[1]{\rotatebox{90}{\parbox{2.4cm}{\raggedright #1}}}

\newcommand{\methodname}[1]{\parbox[c]{2.4cm}{\raggedright #1}}

\begin{table}[t]
\centering
\footnotesize
\setlength{\tabcolsep}{6pt}
\caption{A capability comparison of competing methods. Our work is the only approach that jointly provides an object-decomposed, mesh-accurate reconstruction with photorealistic appearance and a completed background from sparse views. Legend: \cmark~= supported, \pmark~= partial support (e.g., per-edit or runtime only), \xmark~= not supported.}
\label{tab:comparison}
\begin{tabular}{l cccccc}
\toprule
\textbf{Method}
 & \compRotHead{Mesh-accurate}
 & \compRotHead{Photorealistic}
 & \compRotHead{Object-decomposed}
 & \compRotHead{Completed background}
 & \compRotHead{Sparse-view}
 & \compRotHead{Interactive} \\
\midrule
\methodname{2DGS} \cite{Huang242DGS}                                   & \cmark & \cmark & \xmark & \xmark & \xmark & \xmark \\
\methodname{SuGaR} \cite{guedon2024sugar}                              & \cmark & \cmark & \xmark & \xmark & \xmark & \cmark \\ [2pt]
\methodname{Gaussian Opacity Fields} \cite{yu2024gaussian}            & \cmark & \xmark & \xmark & \xmark & \xmark & \xmark \\[6pt]
\methodname{Mesh / Triangle Splatting} \cite{held25meshsplatting, held25trianglesplatting} & \cmark & \cmark & \xmark & \xmark & \xmark & \cmark \\
\midrule
\methodname{Gaussian Grouping} \cite{ye2024gaussian}                   & \xmark & \cmark & \cmark & \pmark & \xmark & \cmark \\
\methodname{VR-GS} \cite{jiang2024vr}                                   & \xmark & \cmark & \cmark & \pmark & \xmark & \cmark \\
\methodname{GScream} \cite{wang2024learning}                           & \xmark & \cmark & \xmark & \pmark & \xmark & \xmark \\
\methodname{InFusion} \cite{liu2026infusion}                           & \xmark & \cmark & \xmark & \pmark & \xmark & \cmark \\
\midrule
\methodname{MAtCha} \cite{guedon2025matcha}                            & \cmark & \cmark & \xmark & \xmark & \cmark & \xmark \\
\methodname{Object-centric 2DGS} \cite{rogge2025object}                & \cmark & \cmark & \pmark & \xmark & \xmark & \cmark \\
\methodname{Split\&Splat} \cite{monchieri2026split}                    & \xmark & \cmark & \cmark & \xmark & \xmark & \cmark \\
\midrule
\textbf{Ours}                                             & \cmark & \cmark & \cmark & \cmark & \cmark & \cmark \\
\bottomrule
\end{tabular}
\end{table}

%% file: sec/3_00_methodology.tex
\section{Methodology}
\label{sec:methodology}

Our method reconstructs a scene not as a single field but as a collection of independently optimized objects and a background, assembled into one environment. The input is a casually captured video of a static scene together with per-frame camera poses; in practice we recover poses with global structure-from-motion~\cite{pan25glomap, Schonberger16COLMAP}, though any posed image set suffices. We build on mesh splatting~\cite{held25meshsplatting}, a surface-splatting representation whose primitives are opaque, spherical-harmonic-shaded triangles initialized from the sparse point cloud. One property of this representation is central to our approach: although primitives may be partially transparent early in optimization, they are driven opaque as training proceeds, so a fitted surface cannot be seen through.

The organizing principle of the pipeline is a partition of image space. Across the posed frames, every pixel is assigned to exactly one component: one of the scene's objects, or the background. Each component is reconstructed independently from the pixels it owns before all components are composed into one scene (Fig.~\ref{fig:pipeline}). Two properties follow directly and motivate the design. First, because the components partition the images (no pixel is supervised twice, and none is left unsupervised), their union covers exactly what a holistic reconstruction would have covered, so the composed scene is free of holes by construction. Second, because each object is a self-contained component in the shared world frame, it remains separable after the fact: it can be selected, moved, or removed without disturbing its neighbors.

\subsection{Scene Decomposition}
\label{subsec:decomposition}
The partition begins with a set of per-frame instance masks that are mutually exclusive: each pixel names at most one object. We obtain these masks in one of two ways, depending on what the capture provides.

When the scene ships with a ground-truth 3D reconstruction carrying per-object annotations, as in ScanNet++ \cite{Yeshwanth23ScannetPP}, we rasterize the annotated geometry into every camera and read off, at each pixel, the object owning the nearest surface. This yields masks with three convenient properties: occlusion is resolved automatically by the depth test, so a chair behind a table simply cedes those pixels to the table; object identity is shared across all frames by construction, so no cross-frame association is needed; and, because each pixel stores a single owner, the masks are mutually exclusive by representation. Their only imperfection is inherited from the source scan (where it is incomplete, a few pixels are left unowned), which the background construction below absorbs. We use this path for our main results, as it isolates the contribution of the reconstruction method from any error in the segmentation.

The second regime targets the practical setting where no such annotation exists. Here we obtain masks with a video segmenter that follows objects through the capture (SAM~3~\cite{carion26sam3}, prompted with the scene's object categories), or with an off-the-shelf video object tracker~\cite{cheng2023tracking, ravi2025sam, carion26sam3} that propagates a set of initial masks across all frames, so that object identity is carried through the video rather than re-inferred and re-associated per frame. This removes the dependence on ground-truth geometry, but such trackers are imperfect in two ways that matter downstream. First, their mask boundaries are not perfectly consistent across views: the same object is delineated slightly differently from frame to frame, and these inconsistencies propagate into the reconstruction. Second, and more damaging, a tracker occasionally merges distinct objects under a single identity (an object together with the surface it rests on, say), so that one object's component is trained on pixels belonging to two scene objects at once. Its reconstruction then fuses their geometry, which breaks separability for interactivity and, because the component is now fit against inconsistent multi-object supervision, degrades mesh and even visual fidelity. We report this regime as a step toward a fully automatic pipeline.

\subsection{Image-Space Partition and Background Completion}
\label{subsec:partition}
Given the masks, we turn each frame into training images for the components that appear in it. For every retained object we write a full-frame image in which only the object's own pixels are kept and all others are masked out. We keep the full frame rather than a tight crop so that each object is optimized against the scene's true cameras; it is the mask, not the framing, that restricts what it learns from. The background is then the per-frame complement of the objects: every pixel not claimed by an object trained in that frame. This makes the partition explicit (each pixel belongs to exactly one component, an object or the background) and is precisely what guarantees the composed scene is gap-free.

Removing the objects, however, leaves the background unobserved wherever an object stood and was never seen from another view. Because these pixels are simply absent from the background's training images rather than supervised as empty, they leave a hole in the reconstructed room only where the capture never revealed the surface behind an object. For interactive use we complete such regions so that no gap is exposed when an object is relocated, using an image inpainter adapted from a pretrained latent diffusion model~\cite{rombach22stableDiffusion}. Following Marigold~\cite{viola2025marigold}, we keep the VAE encoder and decoder frozen and fine-tune only the UNet; we extend its three-channel input with six additional channels, three for the RGB input and three for the hole mask, and generate training data synthetically from indoor imagery~\cite{Roberts21Hypersim} by randomly masking parts of each image. Because the inpainter operates per frame, its completions are not consistent from one view to the next, and training a reconstruction on them would inject that disagreement into the geometry. We therefore use inpainting only to render the final composed scene hole-free for interactivity and qualitative novel-view figures (Fig.~\ref{fig:qualitative_and_interactivity}); it is excluded from all quantitative results, which train the background on its observed pixels alone.

Completion need not stop at the background. Separability introduces a gap of its own: a surface an object rests on, such as a tabletop beneath a keyboard, is reconstructed with a hole where the occluder blocked it, exposed the moment the occluder is moved. We posit that completing such occluded object geometry should further improve reconstruction. Generative amodal completion is the natural tool, but we find that its cross-view inconsistency proves far too damaging for geometry-rich objects and collapses the per-object optimization. To test the premise without that instability, we apply a deterministic proxy on the tractable case of flat supporting surfaces: we recover the occluded region by binary hole-filling of the surface's mask~\cite{soille1999morphological} and inpaint the exposed pixels with Navier--Stokes inpainting~\cite{bertalmio2001navier}. This yields a relatively hole-free supporting surface for interactive use, but we keep it out of our headline metrics: because the ground-truth meshes are themselves incomplete where these surfaces are occluded, a completed surface is scored against a ground truth that lacks it, so the measured F-score drops even where the recovered geometry is plausible. We report this analysis, and the geometry-complete data it would take to evaluate completion properly, in the appendix.

\subsection{Independent Reconstruction and Background Randomization}
\label{subsec:reconstruction}

Each object and the background are reconstructed by the same, unmodified mesh-splatting optimizer, differing only in the masked images they are shown. Every component is optimized against the shared cameras under an identical protocol (the same resolution and iteration budget), and we keep, per component, the checkpoint that scores best on its own held-out views. Because isolated objects converge at very different rates, and far sooner than a holistic reconstruction of the whole scene (refer to appendix for details), this per-component selection matters. Since the components are independent, they are trained in parallel.

Training a surface splat on a masked object, however, exposes a subtlety of the representation that we exploit. Mesh splatting renders a pixel $\mathbf{p}$ by compositing, front to back, the $N$ triangles that cover it~\cite{held25meshsplatting}. Each triangle $t_n$ contributes its spherical-harmonic color $\mathbf{c}_{t_n}$ scaled by its opacity $o_{t_n}$ (the minimum of its three vertex opacities) and a spatial coverage window $I_n(\mathbf{p})\in[0,1]$, attenuated by the transmittance of the triangles in front of it. Compositing the primitives over a background color $\mathbf{b}$ gives

\begin{equation}
C(\mathbf{p}) = \sum_{n=1}^{N} \mathbf{c}_{t_n}\, w_n(\mathbf{p}) \;+\; T(\mathbf{p})\,\mathbf{b}
\label{eq:composite}
\end{equation}

\begin{equation}
w_n(\mathbf{p}) = o_{t_n} I_n(\mathbf{p}) \!\!\prod_{i=1}^{n-1}\!\bigl(1 - o_{t_i} I_i(\mathbf{p})\bigr)
\label{eq:composite2}
\end{equation}

where $T(\mathbf{p}) = \prod_{n=1}^{N}\bigl(1 - o_{t_n} I_n(\mathbf{p})\bigr)$ is the residual transmittance reaching the background. Writing the total foreground weight as $W(\mathbf{p}) = \sum_n w_n(\mathbf{p}) = 1 - T(\mathbf{p})$ and the weight-averaged foreground color as $\bar{\mathbf{c}}(\mathbf{p}) = W(\mathbf{p})^{-1}\sum_n \mathbf{c}_{t_n} w_n(\mathbf{p})$, Eq.~\eqref{eq:composite} is simply $C(\mathbf{p}) = W(\mathbf{p})\,\bar{\mathbf{c}}(\mathbf{p}) + \bigl(1 - W(\mathbf{p})\bigr)\mathbf{b}$. As optimization matures the vertex opacities saturate ($o_{t_n}\!\to\!1$), so a covered pixel becomes opaque and its geometry can no longer be thinned by lowering opacity.

When an object is trained in isolation, every pixel in the masked-out region $\Omega$ should render as empty space, that is, as the background itself. Supervising these pixels toward $\mathbf{b}$, the loss over $\Omega$ is

\begin{equation}
\mathcal{L}_\Omega \;=\; \sum_{\mathbf{p}\in\Omega}\bigl\lVert C(\mathbf{p}) - \mathbf{b}\bigr\rVert^2
\;=\; \sum_{\mathbf{p}\in\Omega} W(\mathbf{p})^2\,\bigl\lVert \bar{\mathbf{c}}(\mathbf{p}) - \mathbf{b}\bigr\rVert^2
\label{eq:maskloss}
\end{equation}

With a fixed background, Eq.~\eqref{eq:maskloss} is minimized trivially by painting the stray primitives the background color ($\bar{\mathbf{c}}=\mathbf{b}$), which leaves constant-colored triangles littering $\Omega$ and corrupting the object. We instead resample $\mathbf{b}$ from a uniform distribution at every iteration. Taking the expectation over $\mathbf{b}$ (mean $\bar{\mathbf{b}}$, per-channel variance $\sigma_\mathbf{b}^2>0$)

\begin{equation}
\mathbb{E}_{\mathbf{b}}\bigl[\mathcal{L}_\Omega\bigr]
\;=\; \sum_{\mathbf{p}\in\Omega} W(\mathbf{p})^2\Bigl(\bigl\lVert \bar{\mathbf{c}}(\mathbf{p}) - \bar{\mathbf{b}}\bigr\rVert^2 + \sigma_\mathbf{b}^2\Bigr)
\label{eq:expmaskloss}
\end{equation}

which, because $\sigma_\mathbf{b}^2>0$ and the colors are bounded, is minimized only as $W(\mathbf{p})\!\to\!0$ for all $\mathbf{p}\in\Omega$: no fixed foreground color can track a background that changes every step. Since the opacities are already saturated and the coverage $I_n(\mathbf{p})$ of a triangle over a pixel it overlaps cannot vanish, $W(\mathbf{p})$ cannot be driven down by making primitives fainter; the only way to satisfy Eq.~\eqref{eq:expmaskloss} is to remove the primitives covering $\Omega$. Background randomization thus turns the masked region into a pressure that deletes every primitive not supported by the object. This requires only setting the renderer's background color $\mathbf{b}$ per iteration, which mesh splatting's differentiable renderer already exposes. It is also what makes the method strong at novel-view synthesis from few captures: a holistic reconstruction sheds such stray, environment-corrupted primitives only once enough viewpoints expose them as inconsistent, whereas background randomization removes them from a single view regardless of how many are available.

\input{tab/sparseview_comparison}

This pruning is also the mechanism behind our fidelity gains, and it clarifies why decomposition helps at all. When a whole scene is fit at once, surfaces that are never directly observed (the underside of a table, the wall behind a cabinet) receive no corrective supervision and are left filled with whatever texture the surrounding geometry happened to deposit, a defect no further training removes because nothing observes it. Reconstructing an object in isolation ameliorates this failure mode: with every other surface masked away and the masked region actively pruned, the optimizer has nothing to fit but the object, and no way to place primitives where the object is not.

Finally, decomposition makes each reconstruction far easier: a single object presents much less geometric and appearance variation than a full room and reaches its best checkpoint in fewer than half the iterations a whole scene requires (refer appendix for details), so few views already constrain it well.
Composition is then trivial. Because every component was fitted in the same world frame against a partition of the same photographs, their union is the scene: we concatenate the per-object and background primitives into a single splat, with no joint re-optimization and no blending. What each component contributes to a rendered pixel is decided by the renderer's ordinary depth ordering, exactly as if the primitives had been optimized together. The result is one scene that renders as a whole, yet in which every object remains an identifiable, separable group of primitives available for downstream selection, editing, or export.

%% file: tab/sparseview_comparison.tex
\begin{table*}[t]
\caption{Comparison of meshsplatting and segmented-meshsplatting across scenes in sparse view (train:test = 1:7)}
\label{tab:sparseview_comparison}
\resizebox{\textwidth}{!}{
\small
\begin{tabular}{llccccccccccc}
  \toprule
  \textbf{Method} & \textbf{Metric}
  & \rotatebox{90}{\textbf{5a269ba6fe}} & \rotatebox{90}{\textbf{08bd80ce2a}} & \rotatebox{90}{\textbf{39f36da05b}}
  & \rotatebox{90}{\textbf{69e5939669}} & \rotatebox{90}{\textbf{cc0aa81452}} & \rotatebox{90}{\textbf{ef18cf0708}}
  & \rotatebox{90}{\textbf{fb564c935d}} & \rotatebox{90}{\textbf{fe5fe0a8a4}} & \rotatebox{90}{\textbf{00a231a370}}
  & \rotatebox{90}{\textbf{08bbbdcc3d}} & \rotatebox{90}{\textbf{Average}}\\
  \midrule
  \multirow{4}{*}{meshsplatting (30k)}
  & SSIM$\uparrow$    & 0.748 & 0.782 & \textbf{0.794} & \textbf{0.820} & 0.618 & 0.687& 0.728 & 0.702 & 0.819 & 0.769 & 0.747 \\
  & PSNR$\uparrow$    & 20.207 & 21.731 & \textbf{23.034} & \textbf{22.422} & 14.100 & 17.656 & 20.297 & 18.868 & 20.751 & 21.362 & 20.043 \\
  & LPIPS$\downarrow$ & 0.367 & 0.310 & \textbf{0.288} & \textbf{0.277} & 0.429 & 0.414 & 0.375 & 0.385 & 0.322 & \textbf{0.322} & 0.349 \\
  & F-score$\uparrow$ & 0.412 & 0.498 & 0.390 & 0.521 & 0.275 & 0.359 & 0.400 & 0.426 & 0.497 & 0.437 & 0.421 \\

  \midrule
  \multirow{4}{*}{meshsplatting (best checkpoint)}
  & SSIM$\uparrow$    & 0.769 & \textbf{0.786} & 0.788 & 0.811 & 0.675 & 0.714 & 0.740 & \textbf{0.717} & 0.841 & \textbf{0.777} & 0.762 \\
  & PSNR$\uparrow$    & 20.545 & \textbf{22.168} & 22.742 & 22.024 & 15.931 & 18.557 & 20.577 & \textbf{19.219} & 21.415 & \textbf{21.421} & \textbf{20.460} \\
  & LPIPS$\downarrow$ & 0.358 & 0.306 & 0.297 & 0.293 & 0.415 & 0.402 & 0.372 & 0.371 & 0.297 & 0.328 & 0.344 \\
  & F-score$\uparrow$ & 0.417 & 0.514 & 0.417 & 0.548 & 0.299 & 0.366 & 0.435 & 0.435 & 0.505 & 0.456 & 0.439 \\

  \midrule
  \multirow{4}{*}{segmented-meshsplatting}
  & SSIM$\uparrow$    & \textbf{0.778} & 0.785 & 0.786 & 0.810 & \textbf{0.687} & \textbf{0.716} & \textbf{0.749} & 0.712 & \textbf{0.847} & 0.759 & \textbf{0.763} \\
  & PSNR$\uparrow$    & \textbf{20.793} & 21.737 & 22.594 & 21.792 & \textbf{16.486} & \textbf{18.768} & \textbf{20.816} & 18.921 & \textbf{21.705} & 20.935 & 20.455 \\
  & LPIPS$\downarrow$ & \textbf{0.339} & \textbf{0.294} & 0.293 & 0.284 & \textbf{0.409} & \textbf{0.398} & \textbf{0.353} & \textbf{0.366} & \textbf{0.283} & 0.332 & \textbf{0.335} \\
  & F-score$\uparrow$ & \textbf{0.487} & \textbf{0.554} & \textbf{0.452} & \textbf{0.581} & \textbf{0.367} & \textbf{0.401} & \textbf{0.468} & \textbf{0.485} & \textbf{0.543} & \textbf{0.491} & \textbf{0.483} \\

  \bottomrule
  \end{tabular}
}
\end{table*}

%% file: sec/4_00_experiments_results.tex
\section{Experiments}
\label{sec:experiments}

\subsection{Setup}
\label{subsec:setup}
\textbf{Datasets.} Our approach needs two things of a dataset: per-frame instance masks consistently associated with the same 3D objects across views, and a ground-truth mesh for measuring geometric fidelity. ScanNet++~\cite{Yeshwanth23ScannetPP} dataset satisfies both: its per-object 3D annotations rasterize into every frame as mutually consistent masks (Sec.~\ref{subsec:decomposition}), and its meshes let us measure F-score directly (scene selection in the appendix). All main results use ScanNet++. We also tried MipNeRF-360~\cite{barron22mipnerf360}, but it has no ground-truth mesh and no instance annotations, and even state-of-the-art video segmentation associates objects inconsistently across its frames, degrading the per-object optimization; we relegate its results to the appendix as evidence the method runs without ground-truth masks.

\noindent\textbf{Novel-view protocol.} To probe reconstruction from few views we invert the usual split: whereas vanilla mesh splatting trains on 7 of every 8 frames and tests on the 8th, we train on 1 and test on the other 7, spreading the single training view evenly through the capture. Every method sees the same 1-in-8 training frames and is scored on the same held-out 7. Because the held-out frames differ from the training view by changes in perspective, this measures novel-view synthesis quality, sharpened by the scarcity of training views.

\noindent\textbf{Metrics.} We report visual fidelity with PSNR, SSIM, and LPIPS~\cite{zhang2018unreasonable}, and geometric fidelity with F-score against the ground-truth scan.

\noindent\textbf{Baselines.} 
Our primary baseline is the original mesh splatting implementation~\cite{held25meshsplatting}, for which we report results from the best-performing checkpoint for each scene (evaluated at both a fixed 30k-iteration, as is common in related literature, and at convergence). A direct quantitative comparison to the closest concurrent works (Sec.~\ref{subsec:rw_segmentation}) was precluded by the absence of public implementations or results.

\noindent\textbf{Implementation.} As a preprocessing step, we use COLMAP~\cite{Schonberger16COLMAP} to undistort the ScanNet++ images. Each scene component is then trained independently for 30k iterations using the unmodified mesh-splatting optimizer. For the final scene composition, we select the best-performing checkpoint for each individual component. Components are independent and trained in parallel across GPUs. Reported metrics use decomposition and background randomization: the background is trained on its observed pixels, and neither the diffusion inpainter nor the deterministic surface completion of Sec.~\ref{subsec:partition} is applied. Both are used only for the qualitative interactivity results and are analyzed in the appendix.

\subsection{Results}
\label{subsec:main_results}

Table~\ref{tab:sparseview_comparison} reports ScanNet++ results across ten scenes. Our object-decomposed reconstruction improves geometric fidelity (F-score) over vanilla mesh splatting on every scene, at both the 30k and best-checkpoint settings, while PSNR and SSIM stay comparable and LPIPS improves on most scenes: decomposition sharpens geometry substantially and appearance modestly. The circular table in Fig.~\ref{fig:qualitative_and_interactivity}(a) also ablates background randomization; removing it introduces contaminated geometry and textures to the underside, confirming that the pruning it induces is a primary source of the geometric gain rather than a byproduct of decomposition alone. Under the conventional dense split (7 train, 1 test) the same comparison shows a 2\% improvement in F1 scores while visual fidelity remains comparable, and the method also runs without ground-truth annotations on MipNeRF-360; both are reported in the appendix.

\subsection{Analysis}
\label{subsec:analysis}
\textbf{Why decomposition helps novel-view synthesis.} The gain comes from how stray, environment-corrupted object primitives are removed. A holistic reconstruction can trim them, but only given enough viewpoints to expose them as inconsistent; as views grow scarce it loses that signal, whereas background randomization prunes them from a single view, and so regardless of view count. Decomposition also eases optimization: each object is a far simpler target and converges in fewer than half the iterations of a full scene (refer to the appendix). We analyze deterministic surface completion, which is excluded from the results above, in the appendix.



\subsection{Interactivity}
\label{subsec:interactivity}
Because every object is a separable group of primitives in a shared world frame, the composed scene supports downstream editing directly. Fig.~\ref{fig:qualitative_and_interactivity} shows objects moved, removed, and rearranged, and the exported meshes brought into a standard game engine. For these qualitative results we fill regions revealed behind relocated objects with the diffusion inpainter of Sec.~\ref{subsec:partition}, making the scene hole-free for interaction; because its completions are not cross-view-consistent, it is used only here and not in our quantitative evaluation (see appendix and Sec.~\ref{sec:limitations}).


%% file: sec/5_00_limitations.tex
\section{Limitations and Future Work}
\label{sec:limitations}

Our largest bottleneck is cross-view consistency. Decomposing a scene and reconstructing its parts independently only works when each part is delineated the same way across every frame, and four modules must each supply that consistency: object segmentation, cross-frame association, and occluded-surface completion, and background inpainting. Each falls short today. Automatic video segmentation and tracking produce boundaries that drift from frame to frame and occasionally merge distinct objects, which is why our headline results fall back on ground-truth masks; even the annotation-rasterized masks of ScanNet++ are not perfectly consistent, as the source scans are incomplete (see appendix). Diffusion background inpainting and generative amodal completion~\cite{wu2025amodal3r} are likewise not multi-view-consistent enough to train a reconstruction on, so our quantitative results exclude both. Surface completion is excluded for a second reason: the ground-truth meshes are themselves incomplete precisely in the occluded regions completion targets, so a completed surface is scored as a false positive against a ground truth that lacks it, and the measured F-score falls even where the recovered geometry is plausible. We therefore cannot presently confirm completion's benefit through geometric metrics, only its qualitative role in producing hole-free interactive scenes. 
The absence of a suitable dataset is itself a limitation on two fronts. No dataset we found supplies consistent per-frame instance masks tied to 3D objects alongside a ground-truth mesh except ScanNet++, which constrains both evaluation and the scenes we can study; and measuring completion fairly would additionally require geometry-complete ground truth, such as realistic synthetic CAD scenes, which do not yet exist for this setting. Building such a dataset is an important direction for making completion-based methods measurable.

A second set of limitations is inherited from the underlying mesh-splatting representation. The recovered meshes are not watertight, so they cannot be driven directly by a physics engine; their textures bake in the capture's lighting, precluding accurate relighting; and their surfaces are spikily and haphazardly triangulated, which costs visual fidelity on close inspection and yields inaccurate contact geometry for interaction. Our reconstructions also assume a static scene and depend on recovered camera poses, and per-object optimization adds cost, though it is largely offset by training components in parallel. Improving the mesh quality of the base representation would compound directly with the fidelity gains from decomposition.

%% file: sec/8_00_conclusion.tex
\section{Conclusion}
\label{sec:conclusion}

We presented a decompose-before-reconstruct approach that reconstructs a scene as independently optimized objects and a background over a partition of image space, instantiated on mesh splatting. Its central ingredient is a background-randomized training scheme that, by exploiting the non-transparency of mesh-splatting primitives, prunes stray geometry during optimization and drives our gains in mesh fidelity and novel-view synthesis, with the largest margins when input views are few. The same decomposition yields a separable, mesh-exportable scene that is directly usable in downstream applications. Our analysis of completion shows that its benefit is presently hard to measure rather than absent: a recovered surface is penalized against ground-truth meshes that are themselves incomplete. The cross-view inconsistency of current generative methods for inpainting, completion, and segmentation remains a primary bottleneck, indicating that future gains will come from developing more consistent models and the geometry-complete datasets needed to train them.


%% file: sec/9_00_appendix.tex
\section{Scene Selection}
\label{sec:scene_selection}
For our experimental evaluation, we select a subset of 10 scenes from the ScanNet++~\cite{Yeshwanth23ScannetPP} dataset. To ensure a rigorous and fair assessment of our decomposition and reconstruction pipeline, these scenes were curated based on the following two primary criteria:
\begin{itemize}
    \item \textbf{Ground-Truth Geometry:} The chosen scenes exhibit high-quality, relatively complete ground truth (GT) meshes, which is critical for accurate quantitative geometric evaluation and generation of complete 2D masks via backprojection.
    \item \textbf{Object Composition:} The environments contain a sufficient number of distinct, decomposable object instances that are highly relevant to our methodology (specifically for ScanNet++: chairs, tables, sofas, coffee tables, benches, PCs, and pool tables).
\end{itemize}

Furthermore, to specifically evaluate the usefulness of OpenCV's \cite{Bradski00OpenCV} Navier-Stokes-based deterministic surface completion, we consider a narrower subset of 4 scenes. Because our heuristic for identifying fillable occlusion holes is intentionally conservative to avoid aggressive geometric hallucinations, we focus on scenes featuring planar surfaces (e.g., desks, tables, etc).




\section{Modality Disparity and Backprojection Mask Incompleteness}
\label{sec:modality_disparity}

\begin{figure}[htbp]
    \centering
    \includegraphics[width=\linewidth]{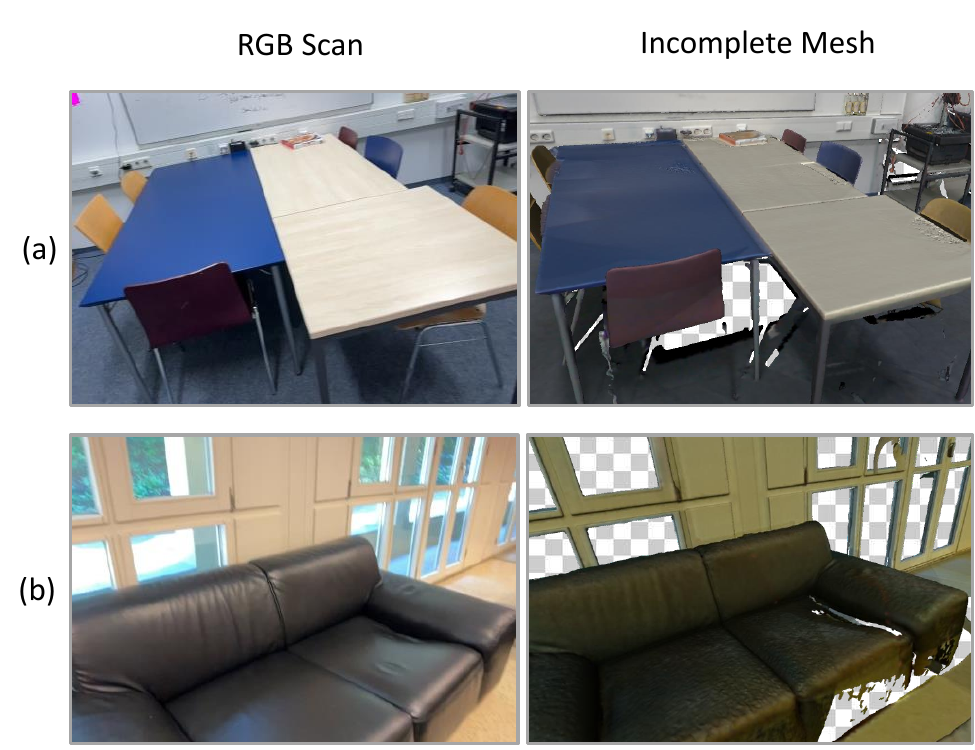}
    \caption{\textbf{Limitations of 3D-to-2D mask projection due to incomplete geometry in ScanNet++.} 
    Due to missing geometry and reconstruction artifacts in the 3D meshes (visible as checkerboard background patterns indicating holes), projecting 3D segmentations to 2D viewports yields incomplete masks and false visibility boundaries. 
    Specific failure cases include: 
    \textbf{(a)} occluded under-desk regions and thin structures such as chair legs, and 
    \textbf{(b)} transparent/specular surfaces (e.g., window panes) as well as dark, non-Lambertian surfaces (e.g., leather sofa folds).}
    \label{fig:mask_inconsistency}
\end{figure}

A key challenge in leveraging 3D annotations for 2D mask generation is the fundamental modality disparity between high-resolution RGB images and sensor-reconstructed 3D meshes. As ScanNet++~\cite{Yeshwanth23ScannetPP} is a real-world dataset, the GT 3D instance annotations are defined on a reconstructed mesh that is inherently imperfect, containing local geometric defects, scanning artifacts, and missing surfaces. While back-projecting these 3D annotations onto the 2D image plane, these underlying mesh defects propagate directly into the 2D domain, illustrated in Fig.~\ref{fig:mask_inconsistency}, yielding incomplete masks that fail to cover the objects entirely. While our method's capacity to improve mesh and visual fidelity in sparse-view settings largely absorbs these degradations, rendering the net impact on our sparse-view reconstruction minor, this modality gap nonetheless imposes an upper bound on performance that a geometrically pristine 3D annotation source would resolve. 

Albeit this mask incompleteness, the GT 3D annotations remain multi-view consistent, which is critical for current splatting methods. Cross-view consistency issues get significantly exacerbated when transitioning to ML-based segmentation pipelines (e.g., using 2D video tracking models such as SAM3 \cite{carion26sam3}). Consequently, we utilize the rasterized ground-truth annotations from ScanNet++ for our primary evaluations.

\begin{figure*}[t]
  \centering
  \includegraphics[width=\textwidth]{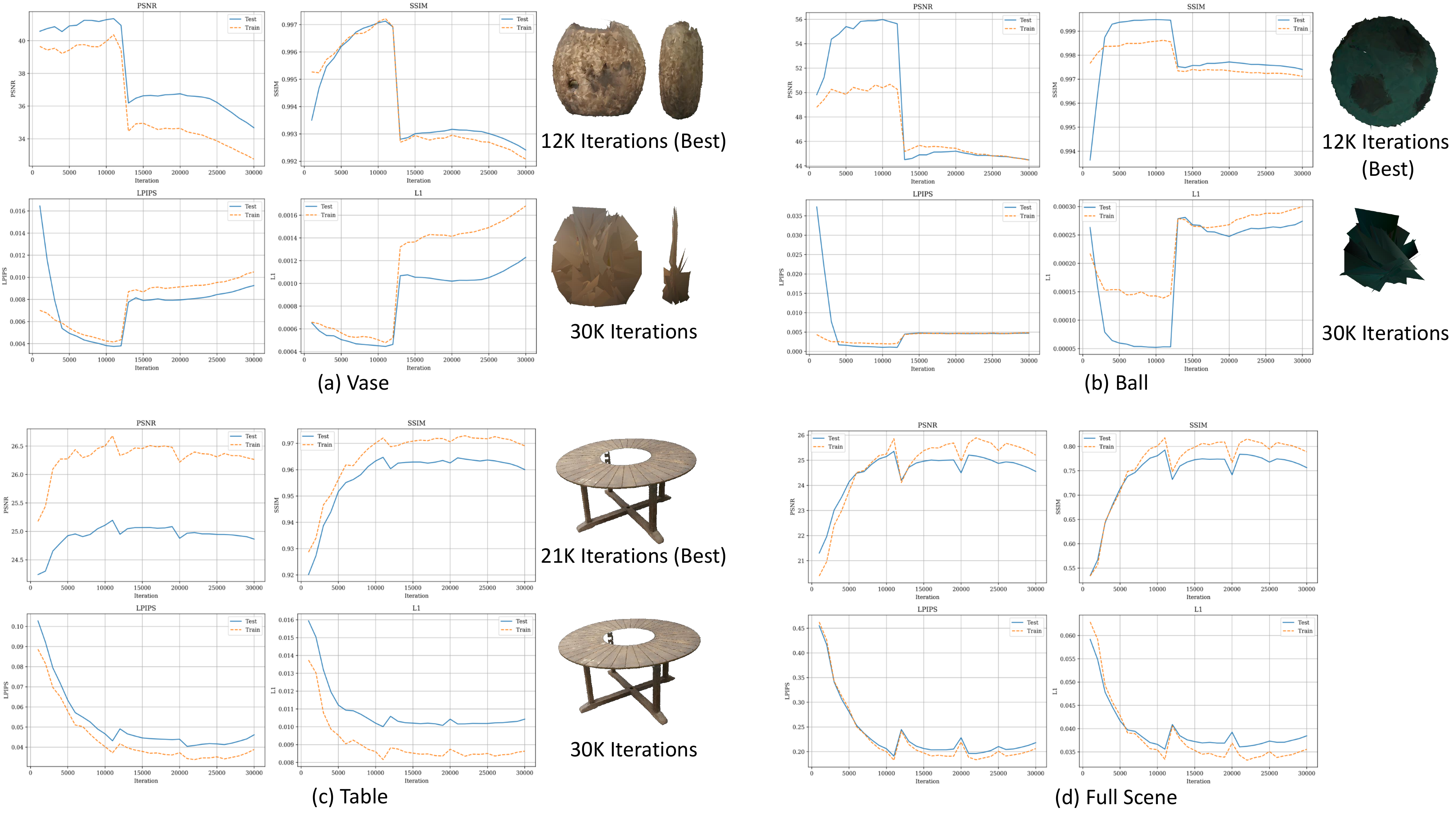}
  \caption{\textbf{Training Graph.} Individual objects reach their best-scoring checkpoints in fewer iterations than the whole-scene reconstruction. Plus, objects converge at markedly different rates from one another primarily depending on their shape simplicity. This figure shows training and validation graphs (PSNR, SSIM, LPIPS, L1=m*PSNR+n*SSIM+o*LPIPS) throughout the 30k optimization iterations.}
  \label{fig:convergence}
\end{figure*}

\section{Per-Component Training Convergence}
\label{sec:convergence}
The variance in training dynamics, observed in Fig.~\ref{fig:convergence}, is heavily governed by the physical scale of the target: smaller object instances possess lower geometric and textural complexity compared to large supporting structures or dense backgrounds, thereby requiring significantly fewer optimization iterations to converge. This disparity in convergence rates makes our per-component training strategy highly advantageous, as it allows us to perform best-checkpoint selection individually for each object, avoiding the sub-optimal compromises and over-fitting typical of uniform holistic training. We note that we select our best checkpoint after the Restricted Delaunay Triangulation (RDT) step (11K-th iteration in Mesh-Splatting \cite{held25meshsplatting}) which is run only once during the entire training routine and improves the connectivity of the triangle soup.

\section{Background Inpainting and Amodal Completion}
\label{sec:inpainting_completion}

\subsection{Network Architecture}

\begin{figure}[htbp]
    \centering
    \includegraphics[width=\linewidth]{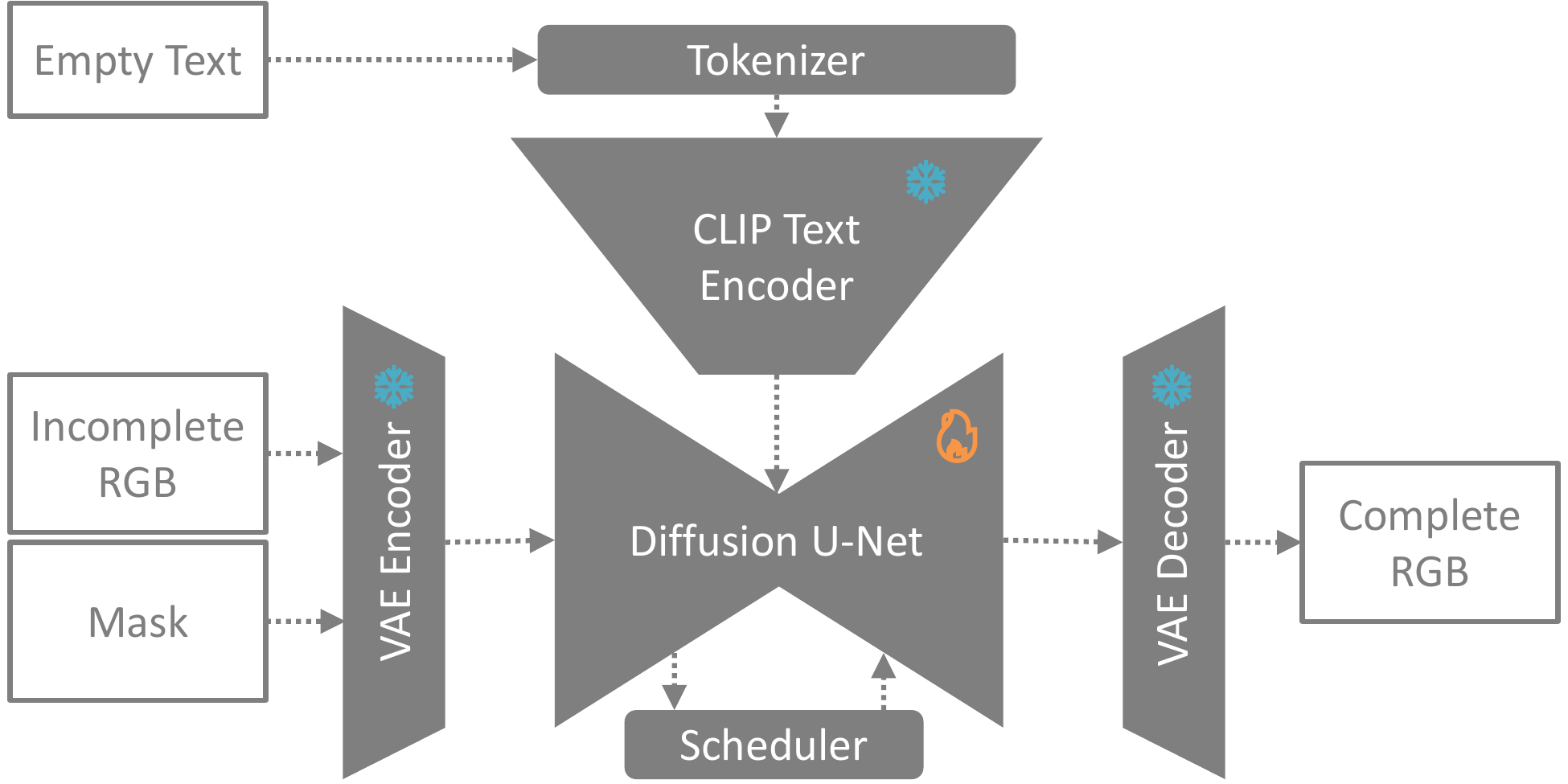}
    \caption{\textbf{Adapted stable diffusion (SD) \cite{rombach22stableDiffusion} architecture for object completion and background inpainting.} We condition on incomplete rgb and mask for the denoising operation, similar to Marigold \cite{viola2025marigold}. Here, we take a pretrained SDv1.5 and modify the input layer of the U-Net. The network is retrained by synthetically generated data from the Hypersim \cite{Roberts21Hypersim} dataset for both background inpainting (Fig. \ref{fig:bg_inpainting}) and object completion (Fig. \ref{fig:completion}); the text encoder and the image encoder-decoder (VAE) are kept frozen. During inference, we employ a variant of the Classifier-Free Guidance approach.}
    \label{fig:diffusion_arch}
\end{figure}

To recover unobserved background regions (inpainting) previously occluded by foreground objects, as well as to generate hidden geometry for partially occluded foreground instances (amodal completion), we design a unified latent diffusion-based architecture. While the structural design of our network takes inspiration from the dense prediction framework introduced in Marigold~\cite{viola2025marigold}, we adapt it for: background inpainting and amodal object completions. 


To leverage generalized image priors, we start with the pre-trained Stable Diffusion v1.5~\cite{rombach2022high}. During the training phase, only the Diffusion U-Net is gets updated. For this retraining, we utilize synthetic indoor environments from the Hypersim dataset~\cite{Roberts21Hypersim}, which provides diverse, photorealistic scenes for learning complex geometries and lighting variations. Though background inpainting and amodal completion are significantly different tasks, we use the same architecture for both models. Background inpainting model is simply trained on randomly masked images. However, generating data for amodal completion is tricky: we use a method similar to \cite{Ozguroglu24pix2gestalt}. All other auxiliary networks, including the VAE encoder, VAE decoder, and text encoders, remain strictly frozen. This selective fine-tuning approach preserves the robust foundational visual priors of the pre-trained latent space, while allowing the U-Net to specialize on our tasks.

\begin{figure}[htbp]
    \centering
    \includegraphics[width=\linewidth]{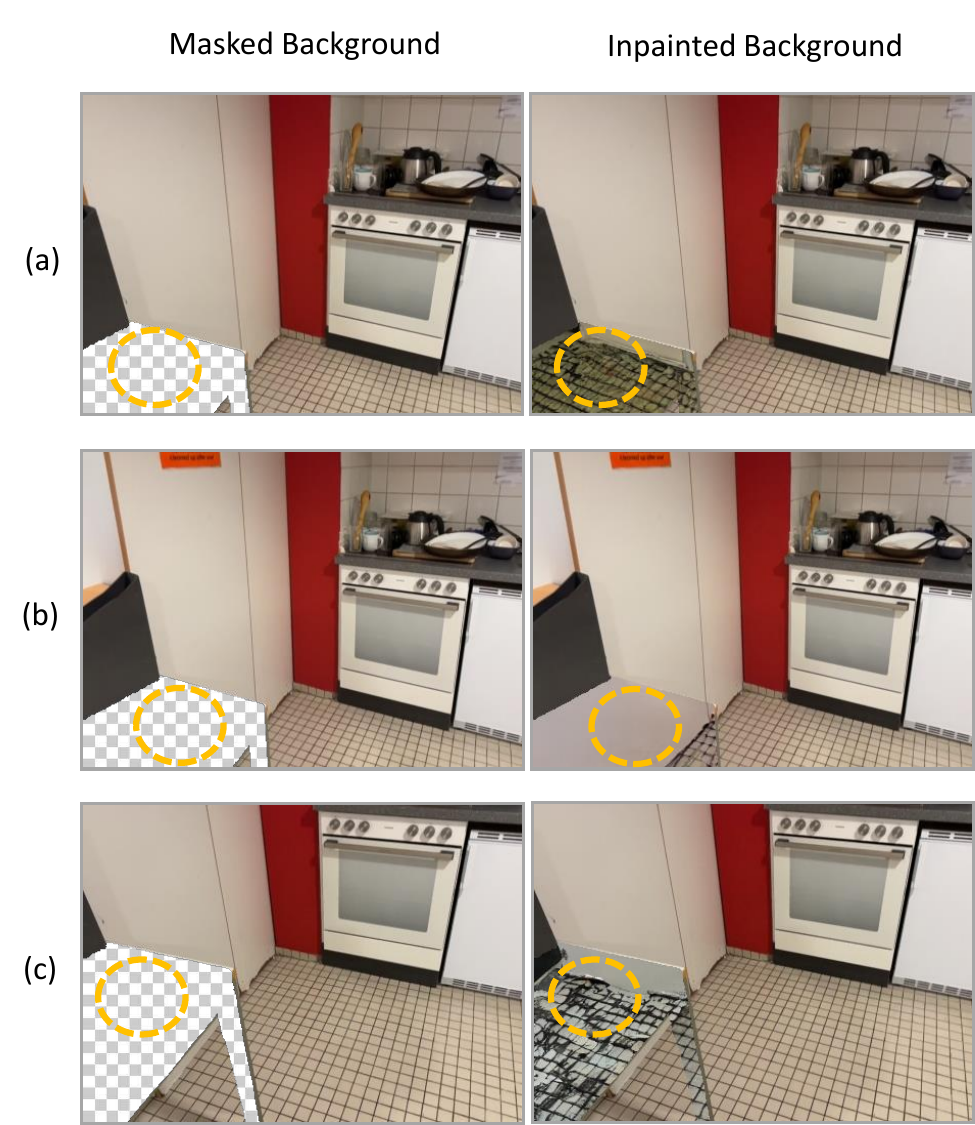}
    \caption{\textbf{Multi-view inconsistency in 2D background inpainting.} While the diffusion model fills the masked regions (left), its independent 2D generation produces inconsistent textures for the same physical area across different views (right). In (a) and (c) the model hallucinates grid patterns but of different color, and (b) has a flat, untextured gray surface. These severe frame-to-frame structural and textural variations create conflicting gradients for splatting methods degrading the 3D mesh optimization.}
    \label{fig:bg_inpainting}
\end{figure}

\subsection{Performance and Optimization Limitations}
\label{subsec:inpainting_performance}

\begin{figure}[htbp]
    \centering
    \includegraphics[width=\linewidth]{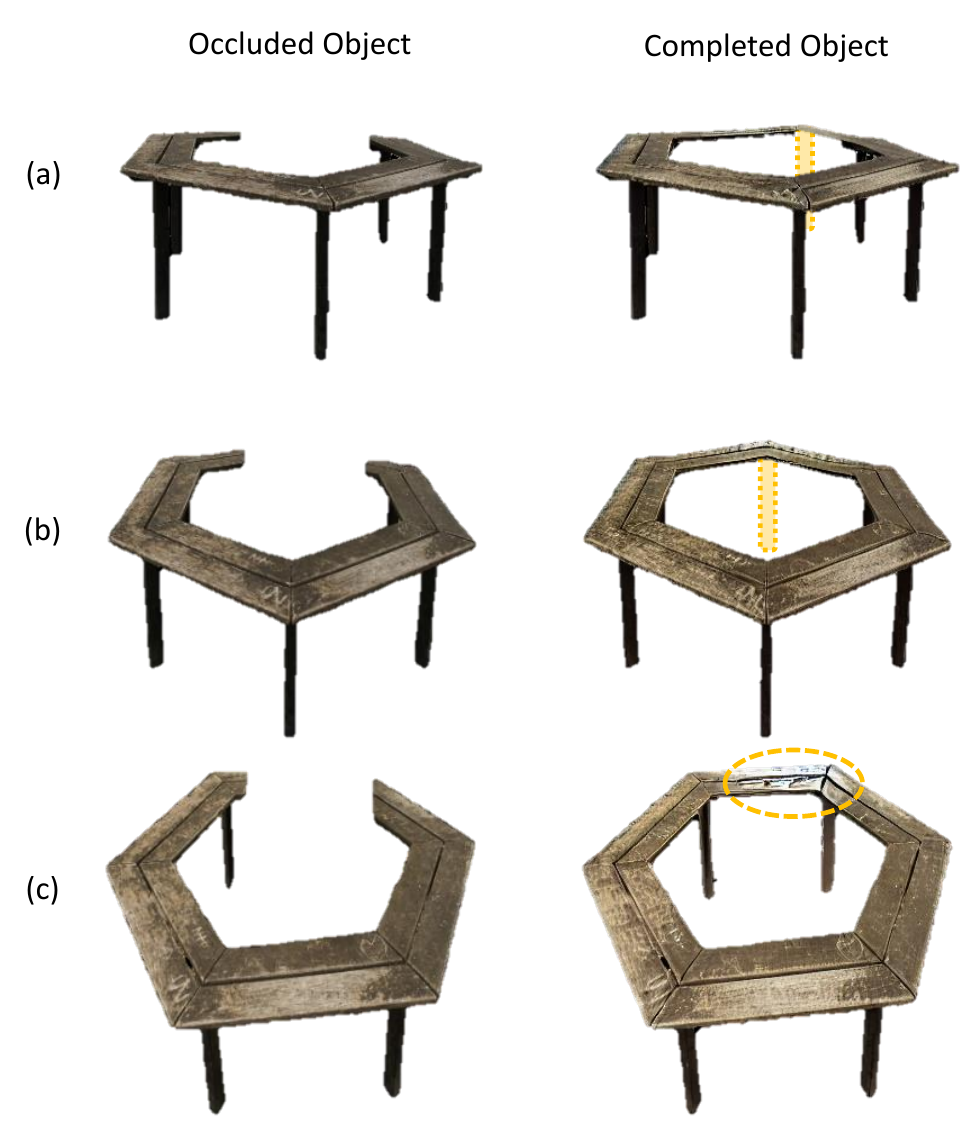}
    \caption{\textbf{Multi-view inconsistency in 2D amodal completion.} While applying independent amodal completion to occluded objects across different frames, the generative model produces inconsistent predictions. (a) and (b) has missing legs, and (c) has severe texture noise and geometric warping. These inconsistencies get propagated and corrupts the splat training.}
    \label{fig:completion}
\end{figure}

While our diffusion-based model is capable of generating visually plausible background and amodal completions for isolated frames, it inherently struggles to maintain strict multi-view consistency. As illustrated for background inpainting in Fig.~\ref{fig:bg_inpainting}, and for object completion in Fig.~\ref{fig:completion}, applying the 2D generative model independently across varying perspectives results in severe spatial inconsistencies. For instance, when generating the occluded regions of a hexagonal wooden table (Fig.~\ref{fig:completion}), the model inconsistently predicts structural elements like missing legs, while introducing severe texture noise and geometric warping across different frames.

Because our 3D splatting optimization relies heavily on photometric consistency across overlapping views, these frame-to-frame generative variations act as contradictory supervisory signals. If injected directly into the optimization pipeline, these conflicting gradients force the model to average over inconsistent appearances, which actively degrades both the reconstructed mesh geometry and the overall texture fidelity. Consequently, while 2D generative completion serves as a valuable tool for downstream interactive scene editing and visualization, both background inpainting and amodal completion are expressly excluded from our optimization pipeline during quantitative evaluations to prevent geometric corruption.

\section{MipNeRF-360 Results}
\label{sec:mipnerf}
\begin{figure}[htbp]
    \centering
    \includegraphics[width=\linewidth]{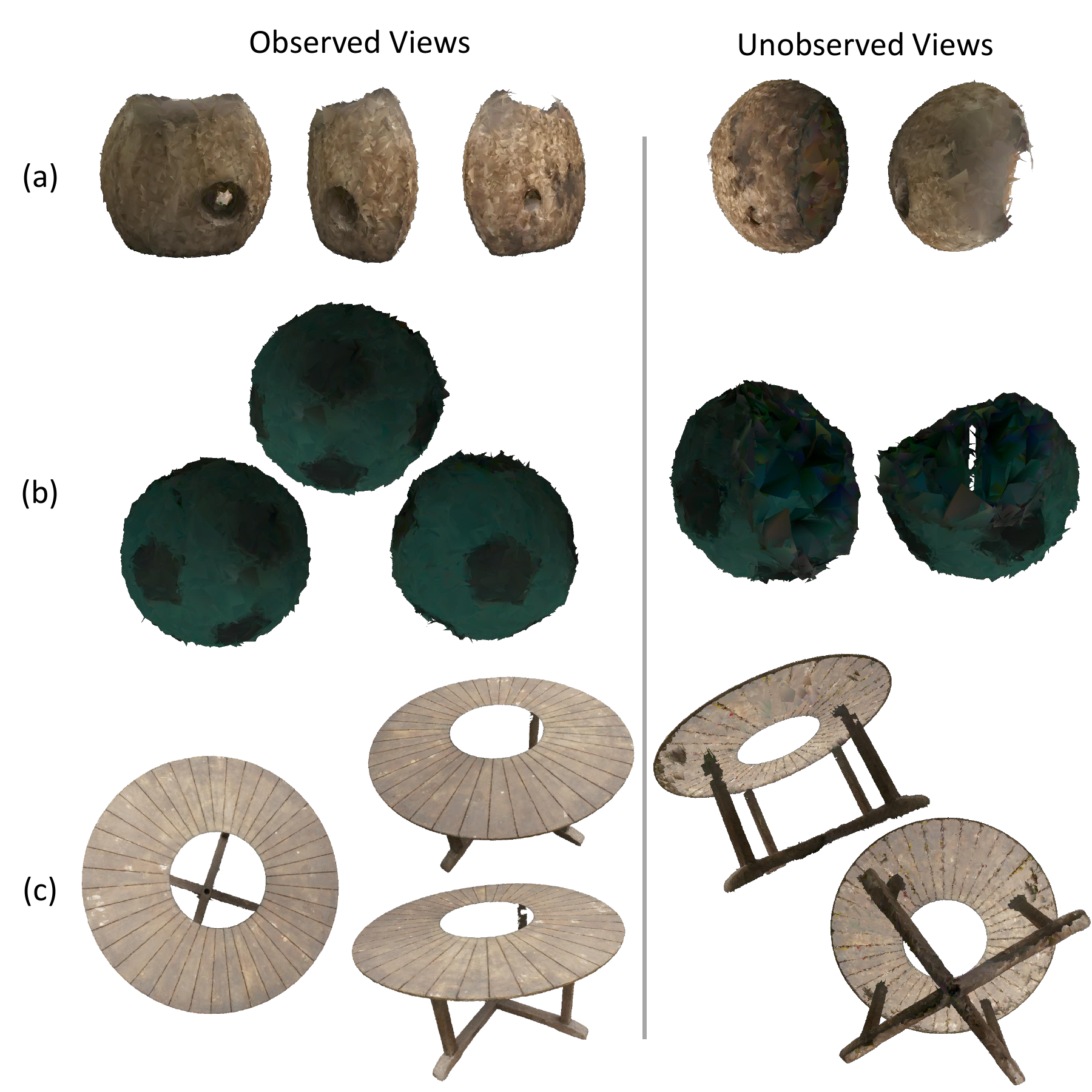}
    \caption{\textbf{Robust mesh and visual reconstruction across extreme perspective changes on Mip-NeRF~360.} 
    Our approach maintains highly consistent geometry and texture quality from both observed views (\emph{left}) and completely unobserved, novel viewpoints (\emph{right}). 
    Despite significant shifts in perspective, the extracted 3D structures remain complete and artifact-free, demonstrating the generalization capability and spatial consistency of the reconstruction across diverse objects.}
    \label{fig:mipnerf-qualitative}
\end{figure}

ScanNet++ supplies both consistent per-object masks and a ground-truth mesh, so it carries our main results. To show that the method still runs without ground-truth annotations, we also report MipNeRF-360~\cite{barron22mipnerf360}, whose masks are obtained through the automatic video-segmentation path of Section Methodology (Scene Decomposition). Because MipNeRF-360 provides no ground-truth masks, the cross-view inconsistency of automatic segmentation causes a marginal loss of visual fidelity relative to the ground-truth-mask setting; with no ground-truth mesh available, we cannot evaluate its effect on geometric fidelity. However, we present qualitative reconstruction results on various objects from the Mip-NeRF~360 dataset in Fig.~\ref{fig:mipnerf-qualitative}.

\section{Surface Completion}
\label{sec:surface_completion}

\begin{figure}[htbp]
    \centering
    \includegraphics[width=\linewidth]{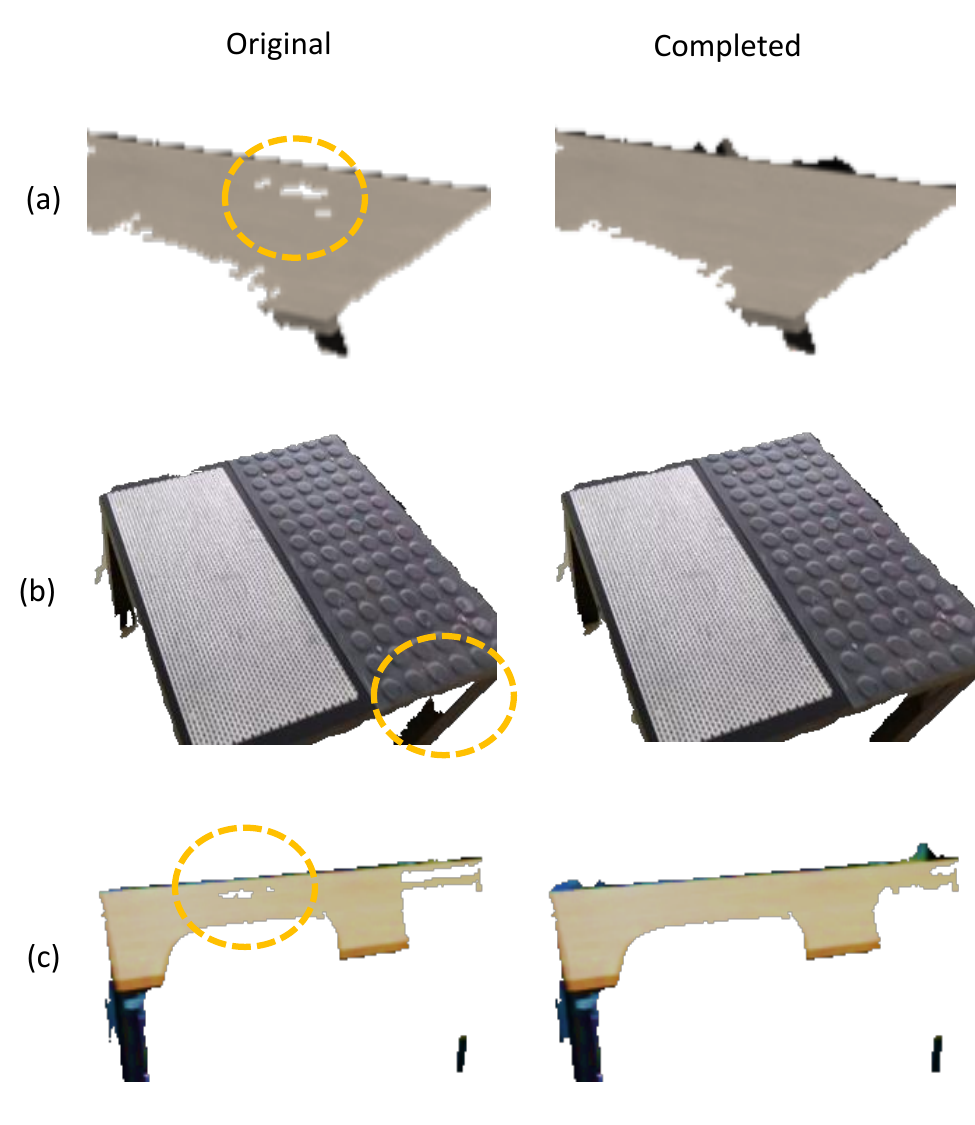}
    \caption{\textbf{Qualitative results of deterministic surface completion.} Original reconstructed supporting surfaces exhibit occlusion holes and broken edges where foreground objects previously stood (left). The OpenCV deterministic completion method successfully repairs these planar regions (right). 
    }
    \label{fig:deterministic_completion}
\end{figure}

The surface completion proposed in our methodology is designed to recover occluded regions of supporting structures (e.g., tabletops, desks) that are shadowed by foreground object instances. Fig. \ref{fig:deterministic_completion} shows results from this method. By patching these occlusions, we reconstruct a relatively complete background mask suitable for downstream interactive editing. To evaluate this module, we conduct a comparative analysis on the completion subset of ScanNet++ (the four scenes containing prominent planar structures; see Sec.~\ref{sec:scene_selection}), comparing our final reconstructions with and without surface completion. For the completion pipeline, we utilize OpenCV's Navier-Stokes-based inpainting method (\texttt{cv2.inpaint}) over the projected planar surfaces.

As shown in Table~\ref{tab:completion_comparison}, executing surface completion results in a marginal decrease in 3D F-score, while leaving 2D rendering metrics (PSNR, SSIM, and LPIPS) virtually unchanged. The invariance of the visual metrics is expected: during evaluation, the segmented foreground objects are rendered in their original positions, thereby re-occluding the completed regions and preventing them from contributing to the image-space loss. We hypothesize that the slight decline in F-score is an artifact of the evaluation target rather than a reflection of geometric degradation. Sensor-based ground-truth meshes (such as those in ScanNet++) are themselves incomplete in occluded regions, meaning that a physically plausible, reconstructed surface is penalized as a false positive when measured against a ground truth that lacks geometry in those coordinates. Fully validating the fidelity of completed surfaces requires a geometry-complete ground truth---such as realistic, synthetic CAD-modeled environments---which does not currently exist at the scale and complexity of ScanNet++. We view the development of such complete synthetic benchmarks as an essential prerequisite for the quantitative evaluation of completion-based reconstruction techniques.

\input{tab/completionVSoriginal}


\input{tab/denseview_comparison}

\section{Dense-View Experiments}

Our primary results focus on sparse views to evaluate our method’s capability for unseen view synthesis. Here, we additionally present results from dense views to demonstrate its scene reconstruction quality, which is on par with vanilla mesh splatting.

In our dense-view experiments (detailed in Table~\ref{tab:denseview_comparison}), we observe that our method performs marginally below holistic mesh splatting. We hypothesize that this performance characteristic is governed by a trade-off between mask quality and view density:
\begin{itemize}
    \item \textbf{Dense-View Regime:} With highly redundant viewpoints, the vanilla mesh splatting baseline can leverage the dense multi-view optimization constraints to effectively prune erroneous geometry on its own. Consequently, the relative margin of improvement from our decomposition vanishes, and the cumulative negative impact of noisy, incomplete boundaries across numerous frames begins to dominate the performance trade-off.
    \item \textbf{Sparse-View Regime:} Conversely, under severe view constraints, the vanilla baseline lacks the redundant perspective supervision necessary to self-prune erroneous geometry, resulting in severe floaters and corrupted surfaces. Under these conditions, the explicit geometric regularization and floater elimination provided by our object-level decomposition heavily outweigh the local errors introduced by imperfect masks, resulting in a substantial net improvement in both mesh and rendering quality without any hyper-parameter tuning.
\end{itemize}

%% file: tab/completionVSoriginal.tex

\begin{table}[t]
\caption{\textbf{Effect of deterministic surface completion.} Quantitative comparison (F-score, PSNR, SSIM, and LPIPS) on 4 selected ScanNet++ scenes, demonstrating the impact of deterministic flat-surface completion.}
\label{tab:completion_comparison}
\centering
\resizebox{\columnwidth}{!}{
\setlength{\tabcolsep}{1pt}
\footnotesize
\begin{tabular}{lcccccccc}
  \toprule
  & \multicolumn{4}{c}{\textbf{without completion}} & \multicolumn{4}{c}{\textbf{with completion}} \\
  \cmidrule(lr){2-5} \cmidrule(lr){6-9}
  \textbf{Scene} & SSIM$\uparrow$ & PSNR$\uparrow$ & LPIPS$\downarrow$ & F1$\uparrow$
  & SSIM$\uparrow$ & PSNR$\uparrow$ & LPIPS$\downarrow$ & F1$\uparrow$ \\
  \midrule
  08bd80ce2a & \textbf{0.785} & 21.737 & \textbf{0.294} & \textbf{0.554} & \textbf{0.785} & \textbf{21.771} & 0.296 & 0.550 \\
  69e5939669 & 0.810 & 21.792 & 0.284 & \textbf{0.581} & \textbf{0.811} & \textbf{21.811} & \textbf{0.283} & 0.580 \\
  fe5fe0a8a4 & \textbf{0.713} & \textbf{18.921} & \textbf{0.366} & \textbf{0.485} & 0.711 & 18.842 & 0.367 & 0.484 \\
  08bbbdcc3d & \textbf{0.759} & \textbf{20.935} & \textbf{0.332} & \textbf{0.491} & 0.755 & 20.708 & 0.338 & \textbf{0.491} \\
  \bottomrule
\end{tabular}
}
\end{table}

%% file: tab/denseview_comparison.tex
\begin{table*}[t]
\caption{\textbf{Dense-view reconstruction results on ScanNet++.} Per-scene and average quantitative metrics (SSIM, PSNR, LPIPS, and F-score) comparing vanilla mesh splatting and our decomposition method under a dense 7:1 train-to-test split.}
\label{tab:denseview_comparison}
\resizebox{\textwidth}{!}{
\small
\begin{tabular}{llccccccccccc}
  \toprule
  \textbf{Method} & \textbf{Metric}
  & \rotatebox{90}{\textbf{5a269ba6fe}} & \rotatebox{90}{\textbf{08bd80ce2a}} & \rotatebox{90}{\textbf{39f36da05b}}
  & \rotatebox{90}{\textbf{69e5939669}} & \rotatebox{90}{\textbf{cc0aa81452}} & \rotatebox{90}{\textbf{ef18cf0708}}
  & \rotatebox{90}{\textbf{fb564c935d}} & \rotatebox{90}{\textbf{fe5fe0a8a4}} & \rotatebox{90}{\textbf{00a231a370}}
  & \rotatebox{90}{\textbf{08bbbdcc3d}} & \rotatebox{90}{\textbf{Average}}\\
  \midrule
  \multirow{4}{*}{meshsplatting (30k)}
  & SSIM$\uparrow$    & 0.849 & \textbf{0.834} & \textbf{0.798} & \textbf{0.846} & \textbf{0.850} & \textbf{0.776} & \textbf{0.794} & \textbf{0.799} & \textbf{0.887} & \textbf{0.869} & \textbf{0.830} \\
  & PSNR$\uparrow$    & \textbf{24.163} & \textbf{25.343} & \textbf{23.874} & \textbf{24.029} & 22.762 & \textbf{22.459} & \textbf{23.389} & \textbf{22.167} & \textbf{24.702} & \textbf{25.563} & \textbf{23.845} \\
  & LPIPS$\downarrow$ & 0.262 & \textbf{0.279} & \textbf{0.305} & \textbf{0.273} & 0.285 & 0.385 & \textbf{0.333} & \textbf{0.319} & 0.256 & \textbf{0.240} & \textbf{0.294} \\
  & F-score$\uparrow$ & 0.466 & \textbf{0.599} & 0.471 & \textbf{0.603} & 0.338 & 0.411 & \textbf{0.509} & \textbf{0.547} & \textbf{0.605} & \textbf{0.560} & \textbf{0.511} \\


  \midrule
  \multirow{4}{*}{Ours}
  & SSIM$\uparrow$    & \textbf{0.852} & 0.802 & 0.790 & 0.812 & \textbf{0.850} & 0.769 & 0.776 & 0.779 & 0.877 & 0.833 & 0.814 \\
  & PSNR$\uparrow$    & 23.704 & 23.200 & 23.482 & 22.261 & \textbf{23.633} & 21.255 & 22.121 & 21.037 & 23.462 & 23.535 & 22.769 \\
  & LPIPS$\downarrow$ & \textbf{0.242} & 0.298 & 0.312 & 0.304 & \textbf{0.279} & \textbf{0.374} & 0.341 & 0.323 & \textbf{0.248} & 0.271 & 0.300 \\
  & F-score$\uparrow$ & \textbf{0.525} & 0.555 & \textbf{0.492} & 0.585 & \textbf{0.412} & 0.466 & 0.492 & 0.458 & 0.502 & 0.547 & 0.503 \\
  \bottomrule
  \end{tabular}
}
\end{table*}